%% file: arxiv.tex
\documentclass{article}

\usepackage[preprint]{neurips_2021}

\usepackage{etoolbox}
\usepackage{subcaption}
\usepackage{booktabs}
\usepackage{multicol}
\usepackage{multirow}
\usepackage{adjustbox}
\usepackage{graphicx}
\usepackage{wrapfig}
\usepackage{csquotes}
\usepackage{float}
\usepackage{authblk}

\makeatletter
\renewcommand\AB@affilsepx{}
\makeatother

\usepackage{hyperref}
\usepackage{url}
\usepackage{color}
\definecolor{Orange}{rgb}{1,0.5,0}
\definecolor{Red}{rgb}{1,0,0}
\definecolor{Green}{rgb}{0,0.8,0.5}
\definecolor{Purple}{rgb}{0.75,0,1}
\definecolor{babypink}{rgb}{0.96, 0.76, 0.76}
\definecolor{azure}{rgb}{0,0.49,1}
\definecolor{periwinkle}{rgb}{0.8, 0.8, 1.0}
\definecolor{Pink}{RGB}{255, 102, 204}

\title{Comparing Human and Machine Bias in Face Recognition}

\author[1]{Samuel Dooley$^*$}
\author[1]{Ryan Downing$^*$}
\author[2]{George Wei$^*$}
\author[3]{\\Nathan Shankar}
\author[4]{Bradon Thymes}
\author[1]{Gudrun Thorkelsdottir}
\author[5]{\\Tiye Kurtz-Miott}
\author[6]{Rachel Mattson}
\author[7]{Olufemi Obiwumi}
\author[1]{\\Valeriia Cherepanova}
\author[1]{Micah Goldblum}
\author[1]{John P Dickerson}
\author[1]{Tom Goldstein}
\affil[1]{University of Maryland\hfill}
\affil[2]{University of Massachusetts Amherst\hfill}
\affil[3]{Pomona College\protect\\}
\affil[4]{\centering{Howard University} \qquad}
\affil[5]{\centering{University of California, San Diego}\protect\\}
\affil[6]{\centering{University of Georgia} \qquad}
\affil[7]{\centering{Haverford College}}
\newtoggle{review}

\newcommand{\authorsident}{[author's identity]}
\newcommand{\reviewing}[2]{\iftoggle{review}{#1}{#2}}

\newcommand{\dataname}{InterRace}
\newcommand{\nsurvey}{545}

\begin{document}

\maketitle

\begin{abstract}
Much recent research has uncovered and discussed serious concerns of bias in facial analysis technologies, finding performance disparities between groups of people based on perceived gender, skin type, lighting condition, etc. These audits are immensely important and successful at measuring algorithmic bias but have two major challenges: the audits (1) use facial recognition datasets which lack quality metadata, like LFW and CelebA, and (2) do not compare their observed algorithmic bias to the biases of their human alternatives. In this paper, we release improvements to the LFW and CelebA datasets which will enable future researchers to obtain measurements of algorithmic bias that are not tainted by major flaws in the dataset (e.g. identical images appearing in both the gallery and test set). We also use these new data to develop a series of challenging facial identification and verification questions that we administered to various algorithms and a large, balanced sample of human reviewers. We find that both computer models and human survey participants perform significantly better at the verification task, generally obtain lower accuracy rates on dark-skinned or female subjects for both tasks, and obtain higher accuracy rates when their demographics match that of the question. Computer models are observed to achieve a higher level of accuracy than the survey participants on both tasks and exhibit bias to similar degrees as the human survey participants.
\end{abstract}

\input{introduction}
\input{priorwork}
\input{dataset/dataset}

\input{experiments}
\input{results}
\input{discussion}

\typeout{}
\bibliographystyle{iclr2022_conference}
\bibliography{bib}

\appendix
\input{appendix}
\input{datasheet}

\end{document}

%% file: introduction.tex
\section{Introduction}

Facial analysis systems have been the topic of intense research for decades, and instantiations of their deployment have been criticized in recent years for their intrusive privacy concerns and differential treatment of various demographic groups. Companies and governments have deployed facial recognition systems~\citep{weise2020,derringer2019surveillance,hartzog2020secretive} which have a wide variety of applications from relatively mundane, e.g., improved search through personal photos~\citep{GooglePhotosFRT}, to rather controversial, e.g., target identification in warzones~\citep{marson2021}.
A flashpoint issue for facial analysis systems is their potential for biased results by demographics~\citep{garvie2016perpetual,lohr2018facial,buolamwini2018gendershades,grother2019face,dooley2021robustness}, which make facial recognition systems controversial for socially important applications, such as use in law enforcement or the criminal justice system.  To make things worse, many studies of machine bias in face recognition use datasets which themselves are imbalanced or riddled with errors, resulting in inaccurate measurements of machine bias.

It is now widely accepted that computers perform as well as or better than humans on a variety of facial recognition tasks~\citep{lu2015surpassing, grother2019face} in terms of {\em accuracy}, but what about {\em bias}? The algorithm's superior overall performance, as well as speed to inference, makes the use of facial recognition technologies widely appealing in many domain areas and comes at enhanced costs to those surveilled, monitored, or targeted by their use~\citep{lewis2019racial,kostka2021between}. Many previous studies which examine and critique these technologies through algorithmic audits do so only up to the point of the algorithm's biases. They stop short of comparing these biases to that of their human alternatives. In this study, we question how the bias of the algorithm compares to human bias in order to fill in one of the largest omissions in the facial recognition bias literature.

We investigate these questions by creating a dataset through extensive hand curation which improves upon previous facial recognition bias auditing datasets, using images from two common facial recognition datasets~\citep{huang2008labeled,liu2015faceattributes} and fixing many of the imbalances and erroneous labels.  Common academic datasets contain many flaws that make them unacceptable for this purpose.  For example, they contain many duplicate image pairs that differ only in their compression scheme or cropping.  As a result, it is quite common for an image to appear in both the gallery and test set when evaluating image models, which distorts accuracy statistics when evaluating on either humans or machines. Standard datasets also contain many incorrect labels and low quality images, the prevalence of which may be unequal across different demographic groups.

We also create a survey instrument that we administer to a sample of non-expert human participants ($n=\nsurvey{}$) and ask machine models (both through academically trained models and commercial APIs) the same survey questions. In comparing the results of these two modalities, we conclude that:
\begin{enumerate}
    \item Humans and academic models both perform better on questions with male subjects,
    \item Humans and academic models both perform better on questions with light-skinned subjects,
    \item Humans perform better on questions where the subject looks like they do, and
    \item Commercial APIs are phenomenally accurate at facial recognition and we could not evaluate any major disparities in their performance across racial or gender lines.
\end{enumerate}
Overall we found that computer systems, while far more accurate than non-expert humans, sometimes have biases that are detectable at a statistically significant level on $t$-tests and logistic regressions.  However, when bias was detected in our studies it was comparable in magnitude to human biases.

%% file: priorwork.tex
\section{Background and Prior Work}

We provide a brief overview of facial recognition and additional related work. We further detail similar comparative studies which contrast the performance of humans and machines. Much of the discussion of bias overlaps with the sub-field of machine learning that focuses on social and societal harms. We refer the reader to~\citet{chouldechova2018frontiers} and~\citet{fairmlbook} for additional background of that broader ecosystem and discussion around bias in machine learning.

\paragraph{Facial Recognition}
In this overview, we focus on a review of the types of facial recognition technology rather than contrasting different implementations thereof. Within facial recognition, there are two large categories of tasks: verification and identification. Verification asks a 1-to-1 question: is the person in the source image the same person as in the target image? Identification asks a 1-to-many question: given the person in the source image, where does the person appear within a gallery composed of many target identities and their associated images, if at all? Modern facial recognition algorithms, such as~\citet{he2016deep,chen2018mobilefacenets,wang2018cosface} and~\citet{deng2019arcface}, use deep neural networks to extract feature representations of faces and then compare those to match individuals. An overview of recent research on these topics can be found in~\citet{wang2018deep}. Other types of facial analysis technology include face detection, gender or age estimation, and facial expression recognition.

\paragraph{Bias in Facial Recognition}
Bias has been studied in facial recognition for the past decade. Early work, like that of~\citet{klare2012face} and~\citet{o2012demographic}, focused on single-demographic effects (specifically, race and gender), whereas the more recent work of~\citet{buolamwini2018gendershades} uncovers unequal performance from an intersectional perspective, specifically between gender and skin tone.  The latter work has been and continues to be hugely impactful both within academia and at the industry level. For example, the 2019 update to NIST FRVT specifically focused on demographic mistreatment from commercial platforms by focusing on performance at the group and subgroup level~\citep{grother2019face}. 

While our work focuses on the identification and comparison of bias, existing work on remedying the ills of socially impactful technology and  unfair systems can be split into three (or, arguably, four~\citep{savani2020posthoc}) focus areas: pre-, in-, and post-processing. Pre-processing work largely focuses on dataset curation and preprocessing~\citep[e.g.,][]{Feldman2015Certifying, ryu2018inclusivefacenet, quadrianto2019discovering, wang2020mitigating}. In-processing often constrains the ML training method or optimization algorithm itself~\citep[e.g.,][]{zafar2017aistats, Zafar2017www, zafar2019jmlr, donini2018empirical, goel2018non,Padala2020achieving, agarwal2018reductions, wang2020mitigating,martinez2020minimax,diana2020convergent,lahoti2020fairness}, or focuses explicitly on so-called fair representation learning~\citep[e.g.,][]{adeli2021representation,dwork2012fairness,zemel13learning,edwards2016censoring,madras2018learning,beutel2017data,wang2019balanced}. Post-processing techniques adjust decisioning at inference time to align with quantitative fairness definitions~\citep[e.g.,][]{hardt2016equality,wang2020fairness}.

\paragraph{Human Performance Comparisons}
No work in the past to our knowledge has specifically focused on the question of comparing bias or disparity between humans and machines. Some prior work has looked at comparing overall performance or accuracy between the two groups. \citet{tang2004face,o2007face,phillips2014comparison} compare human and computer-based face verification performance.  \citet{lu2015surpassing} was the first paper to show machine accuracy outpacing human accuracy. \citet{hu2017person,phillips2018face,robertson2016face} compared face recognition performance of human specific sub-populations 
whereas \citet{white2015error} looked at comparing overall performance of humans who use the \emph{outputs} of face recognition systems. 

%% file: dataset/dataset.tex
\section{\dataname{} Dataset Curation}

We endeavor to answer two research questions: 
{\bf (RQ1) How and to what extent do humans exhibit bias in their accuracy in facial recognition tasks? (RQ2) How does this compare to machine learning-based models?} 
In order to answer these questions, we created a set of challenging identification and verification questions which we posed to humans and machines from a novel dataset called \dataname{} for its application in intersectional facial recognition. The protocol around those experiments are described in Section~\ref{sec:experiments}.

To create our dataset, we first ensured that we had accurately labeled and balanced metadata. This required us to hand-check all the labels in the dataset.  After removing poor quality and redundant images, we found that LFW lacked identities with dark skin tones, which is why further identities were drawn from CelebA. 
Though LFW does have an errata page, CelebA and other facial recognition datasets are known to have many missing or incomplete metadata, and so all CelebA images were examined by an author of this paper before adding them to the dataset. Finally, after randomly generating survey questions, we hand checked that there were no questions for which the answer is apparent or unclear for reasons other than properties of the faces (see Figure~\ref{fig:shortcomings}). In this section we detail our findings about the shortcomings in the metadata labels from LFW and CelebA and outline the steps we took to rectify and supplement these in the creation of the \dataname{} identities.

\subsection{The shortcomings of previous datasets}

\begin{figure}[t]
        \centering
        \begin{subfigure}[t]{.35\textwidth}
            \centering
            \includegraphics[width=.3543\linewidth]{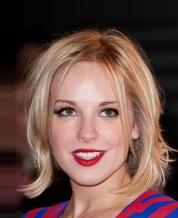}
            \includegraphics[width=.3543\linewidth]{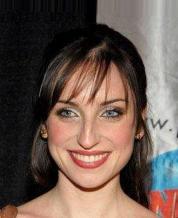}
            \caption{Incorrect Identities: these are labeled as the same; but the left is Zoë Lister and the right is Zoe Lister-Jones. }
            \label{fig:identities}
        \end{subfigure}\hfill
        \begin{subfigure}[t]{.28\textwidth}
            \centering
            \includegraphics[width=.4\linewidth]{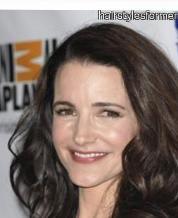}
            \caption{Incorrect Labelling: this individual was labeled as not being pale skinned.}
            \label{fig:labelling}
        \end{subfigure}\hfill
        \begin{subfigure}[t]{.33\textwidth}
            \centering
            \includegraphics[width=.3543\linewidth]{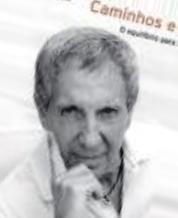}
            \includegraphics[width=.3543\linewidth]{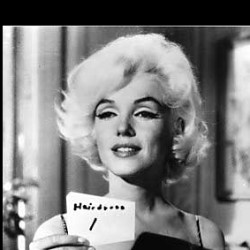}
            \caption{Black and White Images: some identities only have black and white photos.}
            \label{fig:bw}
        \end{subfigure}
        
        \medskip
        
        \begin{subfigure}[t]{.4\textwidth}
            \centering
            \includegraphics[width=.3\linewidth]{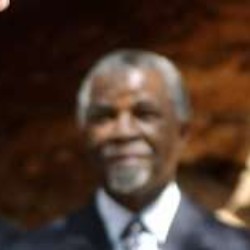}\hfill
            \includegraphics[width=.3\linewidth]{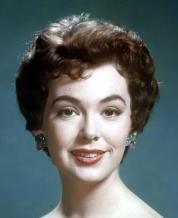}\hfill
            \includegraphics[width=.3\linewidth]{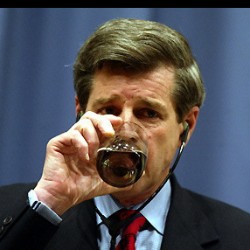}
            \caption{Low-Quality Images}
            \label{fig:qual}
        \end{subfigure}\hfill
        \begin{subfigure}[t]{.35\textwidth}
            \centering
            \includegraphics[width=.4\linewidth]{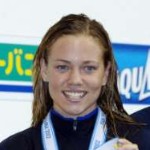}
            \includegraphics[width=.4\linewidth]{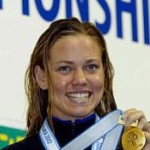}
            \caption{Identical Attire/Background}
            \label{fig:att}
        \end{subfigure}\hfill
        \begin{subfigure}[t]{.25\textwidth}
            \centering
            \includegraphics[width=.5\linewidth]{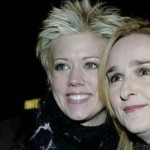}
            \caption{Multiple Distinct Faces}
            \label{fig:mult}
        \end{subfigure}
        
        \caption{Shortcomings present in existing facial identification datasets}
        \label{fig:shortcomings}
    \end{figure}

    In the process of trying to create a reasonable set of identification and verification questions, we identified that the LFW and CelebA datasets generally suffer from a range of problems that distort accuracy and bias metrics. We summarized these problems in~Figure~\ref{fig:shortcomings}.
    
    The first challenge we had to overcome is {\bf incorrect identities}; this includes incorrect names, duplicated identities, as well as clearly incorrect matching between image and name. 
    This problem is particularly harmful for facial recognition models which would be provided with galleries containing incorrect information about identities. 
    In some cases, identities were split across multiple labels due to spellings. We found that this happened almost exclusively with non-canonically western names. E.g., Mesut Ozil (labelled as ``Mesut Zil"), Jithan Ramesh (labelled as ``Githan Ramesh"), Isha Koppikhar (labelled as ``Eesha Koppikhar"), etc. 
    Examples of incorrect identity labels include Neela Rasgotra, a fictional character played by Parminder Singh and ``All That Remains," a band name with the pictured individual being Philip Labonte.
    In other cases, multiple distinct identities were merged into the same label.
    In CelebA, Jennifer Lopez was grouped with Jennifer Driver, and Zoë Lister and Zoe Lister-Jones were both listed under ``Zoe Lister" (pictured in Figure~\ref{fig:identities}). 
        
    Additionally, these datasets exhibit {\bf metadata labelling problems} that manifest in two ways: (1) clearly defined labels being incorrectly or non-uniformly applied, and (2) vague and sometimes harmful metadata. In the first category, CelebA has features such as gender and age which often are incorrect or mislabeled (i.e. a pale-skinned person being labelled as not having pale skin, Figure~\ref{fig:labelling}). Further, many categories in CelebA are subjective and/or harmful. For example, there is a label for ``Attractive," ``Big Nose/Lips," or ``Chubby."

    We found that some identities have {\bf exclusively black and white images} (Figure~\ref{fig:bw}), making it trivial to identity two photos as being of the same label.

    We filtered out {\bf low-quality images} that could not be easily identified for reasons beyond properties of the face, such as poor light exposure, blurriness, facial obstruction, etc. 
    We also removed ``old-timey" photos that were easily associated with a specific time period, as this makes it easy to match them with other similar photos. 
        
    We found that many questions could be answered without considering face features at all, and these were removed. For example if the subject is {\bf wearing identical attire and/or standing in front of an identical background in two images}. Many identities contained multiple images from the same red carpet event or award reception (Figure~\ref{fig:att}).  It {\em very} often happens that the same image appears multiple times in the dataset, but with slightly different crops, compression, or contrast adjustments.  

    Finally, some images {\bf contained multiple faces}. Some of these pictures clearly have one person in the foreground and are therefore not problematic, but in others this is not the case, creating ambiguity as to which person is the target individual. See Figure~\ref{fig:mult}.

    The image types above create inaccuracies when evaluating face recognition systems and distort measurements of bias when these problems occur at rates that differ across groups.  For this reason, many datasets designed for training face analysis systems are not appropriate for evaluating bias. 
    
\subsection{The \dataname{} Identities}
After a thorough review of the LFW and CelebA datasets, random generation of survey questions, and rigorous hand-checking of questions to remove irregularities, we obtained a battery of survey questions for evaluating both humans and machines. We also endeavored to select survey questions that were balanced across gender, age, and skin type. Since LFW is highly skewed towards lighter identities, we included CelebA images and identities as well. We selected identities from LFW with at least two images of an individual, and then we hand labeled each identity for the following: their (1) birth date, (2) country of origin, (3) gender presentation, and (4) Fitzpatrick skin type. Labels 1-3 were assigned by an author of this paper, then that label was checked by at least two other researchers, and modifications were made to achieve agreement among the labelers. Skin type labels (4) were assigned by 8 raters, and the mode was used as the final label.

We note that part of this work does reify categories of gender and skin type that have broader social and political implications. Further, we undertook a task of labeling and categorizing individuals who we do not know and have not received consent from for this task. Every identity for which we created these labels is indeed a celebrity in the public space with Wikipedia entries. Gender labels were rendered from the celebrity's public comments on their own gender identity and/or used pronouns.

The {\bf Fitzpatrick scale}~\citep{fitzpatrick1988validity} was used to help balance the survey to include subjects with diverse skin types. 
This scale is widely used to classify skin complexions into 6 categories. While the Fitzpatrick scale is not perfect, it is the best systematic option currently for ensuring a broadly Representative sample.   

We looked up each celebrity's birth date online, mostly citing Wikipedia, and if we could not find it there, we continued to search on other websites.  However, if we could still not find an individual’s date of birth, we did not list it.  
To find an individual’s {\bf country of origin}, we again cited Wikipedia.  If the individual came from a country that no longer existed (i.e. East and West Germany), we listed the current country.
To label a person’s {\bf gender presentation}, we took note of the person’s preferred pronouns online and in interviews. In the event that their pronouns were not available online, we labeled their gender presentation. A major limitation of the CelebA and LFW datasets is that there were no individuals in our process who identified outside the gender binary or as gender queer.


At the end of our data collection, we collected metadata on 2,545 identities which comprised a total of 7,447 images. The identities themselves are rather imbalanced, though we selected a subgroup from these identities to create a balanced survey, discussed in Section~\ref{sec:experiments}. There are 1,744 lighter-skinned individuals (as defined by Fitzpatrick skin types I-III) and 801 darker-skinned individuals (skin types IV-VI). There are 1,660 males and 885 females. This sample is an improvement over previous datasets as it has been extensively evaluated to remove any errors in labeling and has a robust labeling for a wider array of skin types, unlike previous datasets which chose to label individuals as ``pale.''  
These data have a range of potential future use cases, such as being used for more evaluative facial recognition studies and commercial system audits.

%% file: experiments.tex
\section{Experiments}\label{sec:experiments}

With the high-quality metadata provided in the \dataname{} identities, we conduct two experiments that aim to answer our main research questions regarding the performance disparities of humans and machines.  In this section, we outline how we selected the survey questions, administered the survey to human participants, and evaluated machine models. We describe the results in Section~\ref{sec:results}.

\begin{figure}
    \centering
    \includegraphics[width=.4\linewidth]{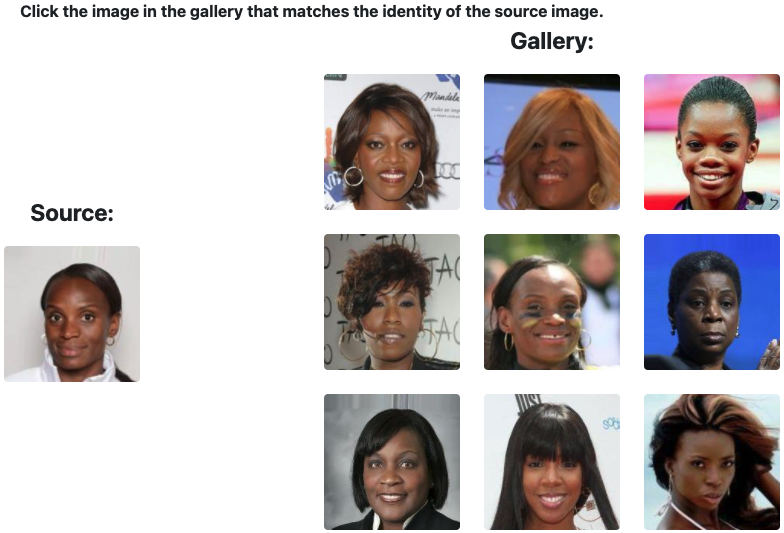}\hfill
    \includegraphics[width=.4\linewidth]{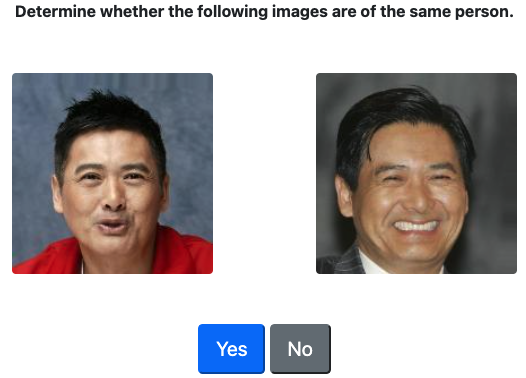}
    \caption{Example questions from the \dataname{} question bank. (Left) An example of an identification question. (Right) An example of a verification question. Notice that the demographics of all identities appearing in a question are matched to ensure the questions are not trivial.}
    \label{fig:queestion_examples}
\end{figure}

For both experiments, we create two types of questions: {\bf identification} and {\bf verification}. Both tasks contain a ``source'' image. In the identification task, 9 other images are presented in a grid, with one being of the same identity as the source and the others being of the same gender and skin type. For the verification task, a second image is selected with equal probability of being the same identity as the source image, or some other of the same gender and skin type as the source.
Examples of these two types of questions can be seen in Figure~\ref{fig:queestion_examples}.

We generated a static question bank with 78 identification questions and 78 verification questions for each of the 12 combinations of gender of skin type.  Of those demographics with more than 78 identities, the source identity for the 78 questions were randomly chosen without replacement. This provided a total of 936 questions for each task.
Finally, a pass was done over all questions to remove any for which context around the face (e.g., background or clothes) could be used to identify a person (e.g., a verification question where both images feature the same sports jersey).
This resulted in a final set of 901 identification questions and 905 verification questions.

\subsection{Human Experiment}\label{sec:human-explain}

\input{table2}

We conducted an institutional review board-approved survey. We collected responses through the crowdsource platform Cint. The survey was split into two parts (whose order was randomized), one for each type of question: identification and verification

Each respondent was asked 36 identification questions and 72 verification questions, for a target survey length of around 10 minutes. The questions for each user were randomly sampled from the total question bank such that an even distribution of questions were asked for each demographic group. As such, each respondent was asked 3 identification questions and 6 verification questions for each intersectional demographic identity. When the user first entered the survey they were prompted with a consent form.
After completing both tasks, respondents filled out a demographic self-identification form which asked the participants their age range, gender, and skin type. When asking respondents to evaluate their own Fitzpatrick skin type scale, we provided a brief description of the scale and respondents were also shown three examples of each skin type from our dataset. The entire text of the survey, including the demographic questions, can be seen in Appendix~\ref{app:survey_qs}.

Within each task, an attention check question was presented after the first five questions and before the last five. For the identification task, the attention check questions used an identical image for the target and in the gallery. For verification, one question consisted of pairing a light skinned female with a dark skin male (obvious negative example), and the other contained two identical images (obvious positive). The images used in these questions do not appear elsewhere in the survey. If a user failed to answer an attention check question correctly, they were screened out and any of their responses were ignored in our analysis. Additionally, any user who passed the attention checks but took fewer than $4$ minutes to complete the survey was dropped from the final analysis. The first $3$ verification and identification questions seen by each user were removed, to account for the possibility that the user may have taken some time to adjust to the format of the questions.

Our survey sampled English-speaking participants who were 18 years or older and were US residents. Our final sample includes \nsurvey{} participants. There are $146$ self-identified as dark-skinned (Fitzpatrick IV-VI) females, $128$ light-skinned (Fitzpatrick I-III) females, $140$ dark-skinned males, and $131$ light-skinned males. Most respondents ($375$) came from the $20-39$ and $40-59$ age demographics. Participants were compensated between \$2.50 and \$5.00 depending on whether the respondent belongs to a part of the population that is harder or easier to reach. Differential incentive amounts, standard in many survey panels~\citep{pewresearchcenter}, were designed to increase panel survey participation among groups that traditionally have low survey response propensities.

\subsection{Machine Experiments}\label{sec:machine-explain}
We conducted experiments with two types of machine models: academic models which we trained ourselves and commercially-deployed models which we evaluated through APIs. Since we do not have to be concerned about question fatigue with machines, we presented all 901 identification and 905 verification questions to the machines.

\paragraph{Academic Models} To measure algorithmic disparities, we trained 6 face recognition models and evaluated them on \dataname{} questions. We trained ResNet-18, ResNet-50 \citep{he2016deep} and MobileFaceNet \citep{chen2018mobilefacenets} neural networks with CosFace \citep{wang2018cosface} and ArcFace \citep{deng2019arcface} losses, which are designed to improve angular separation of the learned features. For the training data, we used images of 9,277 CelebA identities disjoint from identities selected for the \dataname{} dataset. At inference time, the models solve identification questions by finding the closest gallery image in the angular feature space. To solve verification questions, we threshold the cosine similarity between features extracted from images in the pair.

\paragraph{Commercial Models} We evaluated three commercial APIs: AWS Rekognition, Microsoft Azure, and Megvii Face++. We were able to evaluate face verification and identification on AWS and Azure, and only face verification on Face++. 
The AWS CompareFace function, which compares a source and target image, was used for both identification and verification; the target image for identification was one image comprised of the nine gallery images stitched together. Azure has native identification and verification built into their Cognitive Services Face API. Face++ has a similar set up to AWS, however they only compare the largest detected faces in the source and target images; thus we were only able to perform face verification.

\subsection{Analysis Strategy}\label{sec:analysis}
We use a two-tailed $t$-test with matched pairs (with a given pair corresponding to a single respondent's or computer model's scores on the two sections) to compare the accuracy rates between tasks. We also use two-tailed, unpaired $t$-tests to compare the overall accuracy of humans on verification questions with the overall accuracy of computer models on verification questions, and the overall accuracy of humans on identification questions with the overall accuracy of computer models on identification questions. The latter $t$-tests and all $t$-tests referred to in the rest of this section are conducted on the question-level: for instance, when comparing the verification accuracy of humans and machines, we use all verification responses from all human test-takers as one sample, and all verification responses from all machines as the other.

We then analyze the disparity along gender and skin-type categories within our computer algorithms and human survey results. Users and question subjects are binned by skin type. Since the Fitzpatrick is heavily skewed towards Western conceptions of skin tone, we use two categorizations: a binary categorization of ``lighter'' (I-III) and ``darker'' (IV-VI); and categorization by (I-II), (III-IV) and (V-VI). We use two-tailed unpaired $t$-tests to detect the presence of accuracy disparities based on the gender or Fitzpatrick type of the identities that formed the questions. We perform tests of this kind on data from the six individual computer models, and also on the aggregate data sets of all human question responses and all computer algorithm responses.

We use logistic regression in our analysis to allow us to control for confounding variables. Results are reported as odds ratios, which compare the ratio of odds for a baseline event with the odds for a different event. We consider a main model for human subjects which predicts whether an individual question taken by a respondent was answered correctly, with independent variables as the question target gender and skin-type, and test-taker age, gender, and skin-type. The logistic regressions we run on the computer model responses are similar, but do not include test-taker demographics. We do report separate results for different architectures.

%% file: table2.tex
\begin{table}
\centering
\captionof{table}{Demographic breakdown of human survey respondents used in final analysis. }
\footnotesize
\resizebox{.6\linewidth}{!}{
\begin{tabular}{ll|ccccc|c}
\toprule
\toprule
                        & Fitzpatrick & Age  & Age   & Age   & Age   & Age & \multirow{2}{*}{Total}  \\
                        &             & 0-19 & 20-39 & 40-59 & 60-79 & 80+ &                         \\ 
\midrule
\multirow{3}{*}{Male}   & I-II        & 0    & 23    & 37    & 33    & 2   & 95                      \\
                        & III-IV      & 1    & 35    & 18    & 24    & 1   & 79                      \\
                        & V-VI        & 4    & 43    & 33    & 17    & 0   & 97                      \\ 
\midrule
\multirow{3}{*}{Female} & I-II        & 0    & 31    & 26    & 36    & 0   & 93                      \\
                        & III-IV      & 4    & 33    & 26    & 27    & 0   & 90                      \\
                        & V-VI        & 1    & 43    & 27    & 20    & 0   & 91                      \\
\bottomrule
\end{tabular}
}
\label{tbl:demo_resp}
\end{table}

%% file: results.tex
\section{Results}\label{sec:results}

We first provide some overview information about the performance of humans and machines before we move on to answering RQ1 (measuring human bias) and RQ2 (comparing to machine bias). Regression tables can be found in Appendix~\ref{app:tables}.

\paragraph{Verification is Easier Than Identification; Computers are More Accurate Than Humans}
Humans achieved higher accuracy on verification ($78.9\%$) than identification ($68.3\%$, significant with a two-tailed matched-pair $t$-test with $p < 0.001$). For computer models as a whole, this gap persists but is substantially narrowed -- performance on verification is $94.1\%$, with $92.5\%$ on identification ($p = 0.005$). 

The performance difference between machines and humans is highly significant ($p< 0.001$) on both tasks using unpaired $t$-tests which explore group-level changes between the two tasks. Furthermore, even when controlling for demographic effects in a logistic model, humans have a much lower odds compared to computers of getting a question right (OR = $0.23$ for verification, $p < 0.001$, OR = $0.17$ for identification, $p < 0.001$).

\input{table}

\paragraph{Humans and Computers Perform Better on Male Subjects}

For identification questions, we do not observe statistically significant performance gaps for the MobileFaceNet models ($p = 0.3043$ for ArcFace and $p = 0.4752$ for CosFace), but we do observe statistically significant disparities in favor of males for each of the four ResNet models (all $p< 0.04$).
In logistic regression, we observe an odds ratio for computer models on male identification subjects of $1.76$ ($p < 0.001$). Similarly, humans have  significantly ($p<0.001$) better accuracy on identification questions with male subjects: $75.7\%$ on male subjects versus $61.4\%$ on female subjects. The same holds true for humans on verification questions: they attain an accuracy of $81.6\%$ on male subjects, versus $76.1\%$ on female subjects ($p < 0.001$). Interestingly, all demographics of survey respondents (when grouped by gender and skin-type) perform substantially better on males than on females for each task. The results of the human-only logistic models confirm human biases towards male subjects in both verification (OR = $1.39$, $p < 0.001$) and identification (OR = $1.97$, $p < 0.001$). Academic models are found, through logistic regression, to exhibit a statistically significant difference in performance between verification questions with male or female subjects (OR = $1.28$, $p = 0.03$).

\paragraph{Humans and Computers Perform Worse on Darker-Skinned Subjects}

Humans collectively are proportionally $5.2\%$ worse on dark-skinned subjects than light-skinned subjects for verification questions ($80.9\%$ versus $76.7\%$, $p < 0.001$) when we aggregate the Fitzpatrick scale as binary. On identification questions, this proportional difference grew to $11.7\%$ in favor of light-skinned subjects ($72.7\%$ versus $64.2\%$, $p < 0.001$).
This holds even when controlling for the demographics of the respondent: the odds ratio of dark-skinned compared to light-skinned question subjects for verification is  $0.78$ ($p < 0.001$) while for identification it is $0.67$ ($p < 0.001$).
When we aggregate the Fitzpatrick scale as three groups, I-II, III-VI, and V-VI, verification logistic regression finds statistically significant biases in favor of Fitzpatrick types I-II, over both III-VI and V-VI questions compared (OR = 0.93, $p = 0.023$ for III-VI; OR=0.85, $p<0.001$ for V-VI). For the identification task, even when controlling for respondent demographic, question subjects with Fitzpatrick values I-II have higher correct responses than that of values III-VI and V-VI (OR = 0.92, $p = 0.04$ for III-VI; OR=0.70, $p<0.001$ for V-VI).

The results are more nuanced for machines. When we aggregate the Fitzpatrick scale as just ``light'' and ``dark'', we observe a statistically significant proportional disparity of $1.6\%$ in favor of light-skinned question subjects on the verification task ($p = 0.02$); for identification, we do not find evidence of a skin type bias ($p=0.18$).
When we aggregate the Fitzpatrick scale into three categories, I-II, III-IV, and V-VI, we see a disparity for both tasks between the lightest (I-II) and darkest groups (V-VI) ($p=0.004$ and $p=0.04$ for verification and identification respectively).
Academic model performance  is revealed to be significantly different, even when controlling for gender, between the types I-II and V-VI (OR = 0.78, $p = 0.042$ for identification; OR=0.67, $p=0.005$ for verification). However, I-II and III-VI do not show statistically significant differences for academically-trained models (OR = 1.07, $p = 0.591$ for identification; OR=0.93, $p=0.632$ for verification).

\paragraph{Human Test-Takers Perform Better on Subjects of Similar Demographic}

We hypothesized that humans would be more accurate on questions that contained subjects that looked like them. We find evidence to support this hypothesis in our data. On the verification task, humans perform significantly better on questions where the subjects match their gender identity ($1.2\%$, $p = 0.02$), skin type ($1.0\%$, $p = 0.046$), and gender identity and skin type ($1.6\%$, $p = 0.009$). On the identification task, humans perform significantly better on questions where subjects match their skin type ($4.5\%$, $p < 0.001$) and both their gender identity and skin type ($4.7\%$, $p < 0.001$).

\paragraph{Humans and Machines Exhibit Comparable Levels of Disparity}

To test for whether the levels or disparity described above are comparable between humans and machines, we look at the confidence intervals for the odds ratios of comparable models. For both tasks, recall that we observed a disparity on gender and skin type for humans and machines. For verification, we observe that the magnitude of the gender disparities are similar (OR 95\% confidence intervals for humans are [1.33, 1.46] and for academic models are [1.02, 1.61]). For identification, we observe that the magnitude of the gender disparities are also similar (OR 95\% confidence intervals for humans are [1.84,2.10] and for academic models are [1.43,2.17]). As for the skin type disparity, we see similar overlapping confidence intervals between humans and machines for both skin type as binary (light/dark) and ternary (I-II/III-IV/V-VI). This allows us to conclude that when there is a demographic disparity displayed by both humans and machines, the magnitudes and directions of that disparity are statistically similar.

\paragraph{Commercial Facial Recognition Models Are Very Accurate}
The commercial models have very high accuracy, particularly AWS and Face++ which each scored above 97.3\% accuracy on both verification and identification. As a result, these systems do not have enough incorrect responses to have any statistically significant conclusions. On the other hand, Azure achieves verification accuracy of 93.3\% and identification accuracy of 82.9\%. In this case, we see a bias towards question gender in favor of males (OR = 1.76; $p=0.041$) which is comparable to the bias observed with humans and academic models.

%% file: table.tex
\begin{table}
\centering
\captionof{table}{Overall gender and skin type disparities exhibited by the human survey respondents, academic models, and commercial APIs. }
\footnotesize
\resizebox{.65\linewidth}{!}{
\begin{tabular}{lll|cccc}
\toprule
                         &       &  & Human & Academic  & Commercial  \\
                         &       &  &  &  Models &  Models \\\midrule
\multirow{4}{*}{Identification} & \multirow{2}{*}{Darker}   & Female & 55.5\% & 89.9\% & 96.7\%       \\
 &    & Male & 73.1\% & 94.1\% & 97.6\%          \\
 & \multirow{2}{*}{Lighter}   & Female & 67.2\% & 91.3\% & 96.7\%                       \\
 &    & Male & 78.3\% & 94.7\% & 98.7\%                       \\\midrule
\multirow{4}{*}{Verification} & \multirow{2}{*}{Darker}   & Female & 73.4\% & 92.0\% & 97.8\%             \\
 &    & Male & 80.1\% & 94.7\% & 99.9\%                       \\
 & \multirow{2}{*}{Lighter}   & Female & 78.7\% & 94.9\% & 97.6\%                       \\
 &    & Male & 83.1\% & 94.9\% & 98.9\%                       \\\bottomrule
\end{tabular}
}
\label{tbl:metric_comp}
\end{table}

%% file: discussion.tex
\section{Discussion}

The study described in this work is the first to compare disparities and bias between humans and machines. We see that the gender and skin type biases of humans are also present in academic models. Interestingly the level of the disparities present in humans are comparable to that of the machines. These human disparities are present even when controlling for the demographics of the participant. We also find that humans perform better when the demographics of the question match their own. This is not altogether surprising as humans generally spend more time with people of their similar demographics and are more practiced at discriminating faces that look like them. 

One key limitation of our study is that we analyze a crowdsourced sample. While it is demographically diverse, it does not represent a sample of expert facial recognizers. Our results should not be extrapolated too far outside the sample of non-expert crowd workers located in the US. Additionally, the results we have for the computer models are limited to those which we included and do not represent how all models work or behave. 

Our findings contribute meaningfully to the ongoing work of understanding the benefits and harms presented by facial recognition technology. Specifically, we see that automated methods outperform non-expert humans across the board.  When bias is detected in a machine, that bias is comparable to those exhibited by non-expert humans. In the future, further work should examine more targeted populations, such as the direct users of facial recognition technology (e.g., forensic examiners or police officers), to understand how their native bias compares to the biases of machines or human-machine teams.

While our dataset was used here for one specific purpose, we hope that our dataset and survey can be used for future evaluations of the accuracy and bias of facial analysis systems.  Furthermore, we hope our dataset curation process helps bring attention to the many pitfalls and weaknesses of academic datasets.

\paragraph{Ethics Statement}
Our human subjects research was conducted in accordance with the rules, policies, and oversight of our institutional review board (IRB) which deemed our survey collection process to be Exempt. As is common practice with public figures, the data collected was done without the consent of those depicted in the images. This work contributes meaningfully by helping us better understand the tendencies of both humans and machines in this socially important area of facial recognition. The work could potentially be used to improve facial recognition outcomes, concretize the inevitability of facial recognition technology even in morally questionable scenarios, or argue against the future development of facial recognition technologies on the basis of ongoing biases we describe. 

\paragraph{Acknowledgements}
Dooley and Dickerson were supported in part by NSF CAREER Award IIS-1846237, NSF D-ISN Award \#2039862, NSF Award CCF-1852352, NIH R01 Award NLM-013039-01, NIST MSE Award \#20126334, DARPA GARD \#HR00112020007, DoD WHS Award \#HQ003420F0035, ARPA-E Award \#4334192 and a Google Faculty Research Award.  Downing, Wei, Shankar, Thymes, Thorkelsdottir, Kurtz-Miott, Mattson, and Obiwumi were supported by NSF Award CCF-1852352 through the University of Maryland’s REU-CAAR: Combinatorics and Algorithms Applied to Real Problems. We thank Bill Gasarch for his standing commitment to building and maintaining a strong REU program at the University of Maryland.

%% file: appendix.tex
\section{Results Tables}\label{app:tables}

Table~\ref{tbl:human} reports the logistic regressions which depict the bias found between gender and skin type of the subject, even when controlling for respondent demographics. 

Table~\ref{tbl:machine} reports the logistic regressions which depict the bias found between gender and skin type of the subject.

The demographics of the subject in the question are represented with a q (\texttt{qgender} and \texttt{qskin\_type}). The demographics of respondent are represented with an r (\texttt{rgender} and \texttt{rskin\_type}).

\begin{table}[!htbp] \centering 
  \caption{Logistic regressions for {\bf human} performance controlling for gender and skin types (when 2 Fitzpatrick categories are used and when 3 are used)} 
  \label{tbl:human} 
\begin{tabular}{@{\extracolsep{5pt}}lcccc} 
\\[-1.8ex]\hline 
\hline \\[-1.8ex] 
 & \multicolumn{4}{c}{\textit{Dependent variable:}} \\ 
\cline{2-5} 
\\[-1.8ex] & \multicolumn{2}{c}{3 Fitz Categories} & \multicolumn{2}{c}{2 Fitz Categories} \\ 
 & Identification & Verification & Identification & Verification \\ 
\\[-1.8ex] & (1) & (2) & (3) & (4)\\ 
\hline \\[-1.8ex] 
 qgenderMale & 1.965 & 1.394 & 1.973 & 1.395 \\ 
  & t = 20.488$^{***}$ & t = 13.050$^{***}$ & t = 20.556$^{***}$ & t = 13.069$^{***}$ \\ 
  & & & & \\ 
 qskin\_type3III-IV & 0.919 & 0.931 &  &  \\ 
  & t = $-$2.065$^{**}$ & t = $-$2.273$^{**}$ &  &  \\ 
  & & & & \\ 
 qskin\_type3V-VI & 0.697 & 0.846 &  &  \\ 
  & t = $-$9.055$^{***}$ & t = $-$5.378$^{***}$ &  &  \\ 
  & & & & \\ 
 qskin\_type2dark &  &  & 0.667 & 0.779 \\ 
  &  &  & t = $-$12.341$^{***}$ & t = $-$9.846$^{***}$ \\ 
  & & & & \\ 
 rgenderMale & 0.949 & 0.895 & 0.955 & 0.896 \\ 
  & t = $-$1.591 & t = $-$4.386$^{***}$ & t = $-$1.403 & t = $-$4.325$^{***}$ \\ 
  & & & & \\ 
 rskin\_type3III-IV & 1.078 & 1.104 &  &  \\ 
  & t = 1.869$^{*}$ & t = 3.182$^{***}$ &  &  \\ 
  & & & & \\ 
 rskin\_type3V-VI & 1.215 & 1.128 &  &  \\ 
  & t = 4.944$^{***}$ & t = 3.950$^{***}$ &  &  \\ 
  & & & & \\ 
 rskin\_type2dark &  &  & 1.239 & 1.139 \\ 
  &  &  & t = 6.525$^{***}$ & t = 5.119$^{***}$ \\ 
  & & & & \\ 
 Constant & 1.734 & 3.394 & 1.789 & 3.576 \\ 
  & t = 12.960$^{***}$ & t = 36.895$^{***}$ & t = 15.985$^{***}$ & t = 44.590$^{***}$ \\ 
  & & & & \\ 
\hline \\[-1.8ex] 
Observations & 17,877 & 37,605 & 17,877 & 37,605 \\ 
Log Likelihood & $-$10,865.250 & $-$19,292.960 & $-$10,825.090 & $-$19,254.730 \\ 
Akaike Inf. Crit. & 21,744.500 & 38,599.910 & 21,660.190 & 38,519.460 \\ 
\hline 
\hline \\[-1.8ex] 
\textit{Note:}  & \multicolumn{4}{r}{$^{*}$p$<$0.1; $^{**}$p$<$0.05; $^{***}$p$<$0.01} \\ 
\end{tabular} 
\end{table}  

\begin{table}[!tbp] \centering 
  \caption{Logistic regressions for {\bf machine} performance controlling for gender and skin types (when 2 Fitzpatrick categories are used and when 3 are used)} 
  \label{tbl:machine} 
\begin{tabular}{@{\extracolsep{5pt}}lcccc} 
\\[-1.8ex]\hline 
\hline \\[-1.8ex] 
 & \multicolumn{4}{c}{\textit{Dependent variable:}} \\ 
\cline{2-5} 
\\[-1.8ex] & \multicolumn{2}{c}{3 Fitz Categories} & \multicolumn{2}{c}{2 Fitz Categories} \\ 
 & Identification & Verification & Identification & Verification \\ 
\\[-1.8ex] & (1) & (2) & (3) & (4)\\ 
\hline \\[-1.8ex] 
 qgenderMale & 1.763 & 1.279 & 1.762 & 1.282 \\ 
  & t = 5.307$^{***}$ & t = 2.116$^{**}$ & t = 5.304$^{***}$ & t = 2.141$^{**}$ \\ 
  & & & & \\ 
 qskin\_type3III-IV & 1.074 & 0.931 &  &  \\ 
  & t = 0.537 & t = $-$0.479 &  &  \\ 
  & & & & \\ 
 qskin\_type3V-VI & 0.777 & 0.673 &  &  \\ 
  & t = $-$2.038$^{**}$ & t = $-$2.825$^{***}$ &  &  \\ 
  & & & & \\ 
 qskin\_type2dark &  &  & 0.867 & 0.762 \\ 
  &  &  & t = $-$1.371 & t = $-$2.331$^{**}$ \\ 
  & & & & \\ 
 Constant & 10.313 & 16.876 & 10.361 & 16.431 \\ 
  & t = 23.163$^{***}$ & t = 23.625$^{***}$ & t = 27.235$^{***}$ & t = 27.569$^{***}$ \\ 
  & & & & \\ 
\hline \\[-1.8ex] 
Observations & 5,406 & 5,430 & 5,406 & 5,430 \\ 
Log Likelihood & $-$1,423.145 & $-$1,209.350 & $-$1,425.953 & $-$1,211.335 \\ 
Akaike Inf. Crit. & 2,854.291 & 2,426.699 & 2,857.907 & 2,428.671 \\ 
\hline 
\hline \\[-1.8ex] 
\textit{Note:}  & \multicolumn{4}{r}{$^{*}$p$<$0.1; $^{**}$p$<$0.05; $^{***}$p$<$0.01} \\ 
\end{tabular} 
\end{table}

\newpage

\section{Survey Text}\label{app:survey_qs}

In this section, we include the text from the survey described in Section~\ref{sec:human-explain}.

\paragraph{Landing page:}

\begin{displayquote}
Welcome to this survey! It was created at the \reviewing{\authorsident}{ \textbf{Combinatorics and Algorithms for Real Problems (CAAR)} Research Experience for Undergraduates (REU)} during the summer of 2021, made possible by the \reviewing{[author's support]}{University of Maryland College Park and the National Science Foundation (NSF)}.

The survey will take approximately approximately 10 minutes to finish. You will be performing two tasks, with each task taking approximately 5 minutes. After finishing the first task, you will be routed to the other task. You may take a short break in between the two tasks, but the survey is intended to be taken in one sitting. If at any time in the middle of a task you need to take a break, be sure to refresh the page.

Once you feel ready, press `Next' to get routed to your first task.
\end{displayquote}

\paragraph{Verification instructions:}

\begin{displayquote}
Welcome to Task A! This task will take approximately 5 minutes. It has 74 questions. Each question will have two images, each of a single face. Your job is to identify whether the faces in these two images are of the same person or not. You can either click on the buttons `Yes' / `No' or press `y' for `Yes' and `n' for `No'. After you click a button or press one of the `y' or `n' keys, you will not be allowed to change your answer, so keep that in mind. Please try to verify whether the two images are of the same person to the best of your ability. If at any time in the middle of a task you need to take a break, be sure to refresh the page. After finishing the last question, you will be directed to Task B. Once you feel ready, click the `Next' button to start this task.
\end{displayquote}

\paragraph{Verification task heading:}

\begin{displayquote}
\textbf{\textit{\underline{Task A:}}} \textbf{Determine whether the following images are of the same person.}
\end{displayquote}

The ``\textbf{\textit{\underline{Task A:}}}" is a link to a popup that displays the verification instructions again.

\paragraph{Identification instructions:}

\begin{displayquote}
Welcome to Task B! This task will take approximately 5 minutes. It has 38 questions. Each question will have ten images, each of a single face. One image will appear on the left of your screen — this is the target image. The other nine images will appear on the right of your screen in a 3-by-3 grid. Exactly one of these nine images will match the identity of the target image. Your job is to click the image in the grid that matches the target. After you click a picture in the gallery, you will not be allowed to change your answer, so keep that in mind. Please try to identify the matching image to the best of your ability. If at any time in the middle of a task you need to take a break, be sure to refresh the page. After finishing the last question, there will be a brief questionnaire asking about your individual information. Once you feel ready, click the `Next' button to start this task.
\end{displayquote}

\paragraph{Identification task heading:}

\begin{displayquote}
\textbf{\textit{\underline{Task B:}}} \textbf{Click the image in the gallery that matches the identity of the target image.}
\end{displayquote}

Similarly, ``\textbf{\textit{\underline{Task B:}}}" is a link to a popup that displays the verification instructions again.

Note that there is a 50-50 chance for starting on verification or identification. In this case, verification was presented first (which is referred to by ``task A") and identification was presented second (which is referred to by ``task B").

\paragraph{User information page:}

\begin{displayquote}
Please enter your information.

All information will be kept strictly private on secure university servers and will be erased after the completion of this study.

Select your age: [0-19, 20-39, 40-59, 60-79, 80+, Prefer not to say]

Select your gender: [Male, Female, Other]

Feel free to elaborate on your gender presentation: [text-box]

Please select the category that best represents your skin tone.

\underline{What are the Fitzpatrick Skin Types?}

[Pale White Skin, White Skin, Light Brown Skin, Moderate Brown Skin, Dark Brown Skin, Deeply Pigmented Dark Brown Skin]
\end{displayquote}

\paragraph{\underline{What are the Fitzpatrick Skin Types?} popup:}

\begin{displayquote}
The \textbf{Fitzpatrick Skin Phototypes} were developed by dermatologist Thomas B. Fitzpatrick. It is a system commonly used to classify skin complexions and their various reactions to exposure to ultraviolet radiation, or sun exposure. There are 6 catogeries, ranging from extremely sensitive skin which always burns instead of tannning, to very resistant skin which is deeply pigmented and almost never burns.
\end{displayquote}

\paragraph{Thanks page:}

\begin{displayquote}
\textbf{Thanks for taking the \reviewing{\authorsident}{REU-CAAR} 2021 Survey!}

Thanks again for finishing the \textbf{\reviewing{\authorsident}{REU-CAAR} 2021} survey! Have a great rest of your day!
\end{displayquote}

%% file: datasheet.tex
\section{Datasheets for datasets}


\newcommand{\dssectionheader}[1]{%
   \noindent\subsection{%
      {\textbf{{#1}}}
   }
}

\newcommand{\dsquestion}[1]{%
    {\noindent {\textbf{#1}}}
}

\newcommand{\dsquestionex}[2]{%
    {\noindent {\textbf{#1} #2}}
}

\newcommand{\dsanswer}[1]{%
   {\noindent #1 \medskip}
}

\dssectionheader{Motivation}

\dsquestionex{For what purpose was the dataset created?}{Was there a specific task in mind? Was there a specific gap that needed to be filled? Please provide a description.}

\dsanswer{The main purpose for creating this dataset was to create a set of challenging face verification and face identification questions which were used in a series of experiments which compared the bias of humans in machines in these tasks.}

\dsquestion{Who created this dataset (e.g., which team, research group) and on behalf of which entity (e.g., company, institution, organization)?}

\dsanswer{This dataset was created by \reviewing{\authorsident}{8 REU students at Combinatorics and Algorithms for Real Problems (CAAR) under the supervision of John P Dickerson and Tom Goldstein at the University of Maryland, College Park}.}

\dsquestionex{Who funded the creation of the dataset?}{If there is an associated grant, please provide the name of the grantor and the grant name and number.}

\dsanswer{N/A.}

\dsquestion{Any other comments?}

\dsanswer{No.}

\bigskip
\dssectionheader{Composition}

\dsquestionex{What do the instances that comprise the dataset represent (e.g., documents, photos, people, countries)?}{Are there multiple types of instances (e.g., movies, users, and ratings; people and interactions between them; nodes and edges)? Please provide a description.}

\dsanswer{Each instance is an identity. Each identity can have one or more images picturing them.}

\dsquestion{How many instances are there in total (of each type, if appropriate)?}

\dsanswer{There are 2545 unique identities with 7447 images in total.
\begin{table}[H]
\begin{tabular}{|l|l|l|l|}
\hline
\textbf{Gender} & \textbf{Skin Tone} & \textbf{Identities} & \textbf{Images} \\ \hline
Female          & 1                  & 111                 & 236             \\ \hline
Female          & 2                  & 269                 & 760             \\ \hline
Female          & 3                  & 150                 & 443             \\ \hline
Female          & 4                  & 139                 & 303             \\ \hline
Female          & 5                  & 138                 & 301             \\ \hline
Female          & 6                  & 78                  & 156             \\ \hline
Male            & 1                  & 126                 & 250             \\ \hline
Male            & 2                  & 647                 & 2284            \\ \hline
Male            & 3                  & 441                 & 1668            \\ \hline
Male            & 4                  & 189                 & 488             \\ \hline
Male            & 5                  & 171                 & 382             \\ \hline
Male            & 6                  & 86                  & 176             \\ \hline
\end{tabular}
\end{table}
}

\dsquestionex{Does the dataset contain all possible instances or is it a sample (not necessarily random) of instances from a larger set?}{If the dataset is a sample, then what is the larger set? Is the sample representative of the larger set (e.g., geographic coverage)? If so, please describe how this representativeness was validated/verified. If it is not representative of the larger set, please describe why not (e.g., to cover a more diverse range of instances, because instances were withheld or unavailable).}

\dsanswer{This dataset is a sample from both Labelled Faces in the Wild and CelebA. This sample is not representative of the larger set in order to cover a more diverse range of perceived gender and Fitzpatrick rating pairs.}

\dsquestionex{What data does each instance consist of? “Raw” data (e.g., unprocessed text or images) or features?}{In either case, please provide a description.}

\dsanswer{Each instance consists of one of more images of an identity.}

\dsquestionex{Is there a label or target associated with each instance?}{If so, please provide a description.}

\dsanswer{Identity name, approximate age upon release of containing dataset, perceived gender, country of origin, and Fitzpatrick skin rating is labelled for each instance.}

\dsquestionex{Is any information missing from individual instances?}{If so, please provide a description, explaining why this information is missing (e.g., because it was unavailable). This does not include intentionally removed information, but might include, e.g., redacted text.}

\dsanswer{No.}

\dsquestionex{Are relationships between individual instances made explicit (e.g., users’ movie ratings, social network links)?}{If so, please describe how these relationships are made explicit.}

\dsanswer{No.}

\dsquestionex{Are there recommended data splits (e.g., training, development/validation, testing)?}{If so, please provide a description of these splits, explaining the rationale behind them.}

\dsanswer{No.}

\dsquestionex{Are there any errors, sources of noise, or redundancies in the dataset?}{If so, please provide a description.}

\dsanswer{Not that we are aware of.}

\dsquestionex{Is the dataset self-contained, or does it link to or otherwise rely on external resources (e.g., websites, tweets, other datasets)?}{If it links to or relies on external resources, a) are there guarantees that they will exist, and remain constant, over time; b) are there official archival versions of the complete dataset (i.e., including the external resources as they existed at the time the dataset was created); c) are there any restrictions (e.g., licenses, fees) associated with any of the external resources that might apply to a future user? Please provide descriptions of all external resources and any restrictions associated with them, as well as links or other access points, as appropriate.}

\dsanswer{The dataset contains references to images and identities in the Labelled Faces in the Wild and CelebA datasets, whose websites have persistent data catalogs. LFW has an errata page which indicates any errors or updates, but there have been none recently. 

\url{http://vis-www.cs.umass.edu/lfw/}

\url{http://mmlab.ie.cuhk.edu.hk/projects/CelebA.html}}

\dsquestionex{Does the dataset contain data that might be considered confidential (e.g., data that is protected by legal privilege or by doctor-patient confidentiality, data that includes the content of individuals non-public communications)?}{If so, please provide a description.}

\dsanswer{No.}

\dsquestionex{Does the dataset contain data that, if viewed directly, might be offensive, insulting, threatening, or might otherwise cause anxiety?}{If so, please describe why.}

\dsanswer{No.}

\dsquestionex{Does the dataset relate to people?}{If not, you may skip the remaining questions in this section.}

\dsanswer{Yes.}

\dsquestionex{Does the dataset identify any subpopulations (e.g., by age, gender)?}{If so, please describe how these subpopulations are identified and provide a description of their respective distributions within the dataset.}

\dsanswer{The dataset does identify subpopulations, specifically by age, perceived gender, country of origin, and Fitzpatrick skin tones. Age was calculated as the difference between the min of year of death and the release of the containing dataset with the date of birth. Perceived gender was gathered from preferred pronouns. Country of origin was obtained from Wikipedia or another celebrity information page if not available. Fitzpatrick skin tone ratings were determined by at least a 5/8 majority among the \reviewing{\authorsident}{8 REU students}.}

\dsquestionex{Is it possible to identify individuals (i.e., one or more natural persons), either directly or indirectly (i.e., in combination with other data) from the dataset?}{If so, please describe how.}

\dsanswer{Yes since the names are used as the identifiers.}

\dsquestionex{Does the dataset contain data that might be considered sensitive in any way (e.g., data that reveals racial or ethnic origins, sexual orientations, religious beliefs, political opinions or union memberships, or locations; financial or health data; biometric or genetic data; forms of government identification, such as social security numbers; criminal history)?}{If so, please provide a description.}

\dsanswer{This dataset does not contain sensitive information to our knowledge.}

\dsquestion{Any other comments?}

\dsanswer{No.}

\bigskip
\dssectionheader{Collection Process}

\dsquestionex{How was the data associated with each instance acquired?}{Was the data directly observable (e.g., raw text, movie ratings), reported by subjects (e.g., survey responses), or indirectly inferred/derived from other data (e.g., part-of-speech tags, model-based guesses for age or language)? If data was reported by subjects or indirectly inferred/derived from other data, was the data validated/verified? If so, please describe how.}

\dsanswer{The date of birth, perceived gender, and country of origin for a given instance was acquired by manually searching up the given identity's Wikipedia page or any celebrity information page. The Fitzpatrick skin ratings was obtained in a majority vote among the \reviewing{\authorsident}{8 REU students}.}

\dsquestionex{What mechanisms or procedures were used to collect the data (e.g., hardware apparatus or sensor, manual human curation, software program, software API)?}{How were these mechanisms or procedures validated?}

\dsanswer{Manual human curation was used to collect the data.}

\dsquestion{If the dataset is a sample from a larger set, what was the sampling strategy (e.g., deterministic, probabilistic with specific sampling probabilities)?}

\dsanswer{Identities were sampled from the larger LFW and CelebA dataset. Identities from LFW with more than one image were primarily taken. Identities from CelebA were sampled to improve the underrepresented intersectional demographic identities (such as those with Fitzpatrick ratings of I or IV-VI) with a goal of bringing each intersection to above 75 identities.}

\dsquestion{Who was involved in the data collection process (e.g., students, crowdworkers, contractors) and how were they compensated (e.g., how much were crowdworkers paid)?}

\dsanswer{\reviewing{\authorsident}{8 REU students} were involved in the data collection sample and they were compensated as a part of \reviewing{\authorsident}{a Research Experience for Undergraduates program}.}

\dsquestionex{Over what timeframe was the data collected? Does this timeframe match the creation timeframe of the data associated with the instances (e.g., recent crawl of old news articles)?}{If not, please describe the timeframe in which the data associated with the instances was created.}

\dsanswer{The data was collected over a timeframe of two months.}

\dsquestionex{Were any ethical review processes conducted (e.g., by an institutional review board)?}{If so, please provide a description of these review processes, including the outcomes, as well as a link or other access point to any supporting documentation.}

\dsanswer{There was no IRB review conducted for this dataset collection.}

\dsquestionex{Does the dataset relate to people?}{If not, you may skip the remaining questions in this section.}

\dsanswer{Yes.}

\dsquestion{Did you collect the data from the individuals in question directly, or obtain it via third parties or other sources (e.g., websites)?}

\dsanswer{We obtained the data from the individuals through third parties.}

\dsquestionex{Were the individuals in question notified about the data collection?}{If so, please describe (or show with screenshots or other information) how notice was provided, and provide a link or other access point to, or otherwise reproduce, the exact language of the notification itself.}

\dsanswer{Individuals were not notified about the data collection.}

\dsquestionex{Did the individuals in question consent to the collection and use of their data?}{If so, please describe (or show with screenshots or other information) how consent was requested and provided, and provide a link or other access point to, or otherwise reproduce, the exact language to which the individuals consented.}

\dsanswer{No.}

\dsquestionex{If consent was obtained, were the consenting individuals provided with a mechanism to revoke their consent in the future or for certain uses?}{If so, please provide a description, as well as a link or other access point to the mechanism (if appropriate).}

\dsanswer{N/A}

\dsquestionex{Has an analysis of the potential impact of the dataset and its use on data subjects (e.g., a data protection impact analysis) been conducted?}{If so, please provide a description of this analysis, including the outcomes, as well as a link or other access point to any supporting documentation.}

\dsanswer{No.}

\dsquestion{Any other comments?}

\dsanswer{No.}

\bigskip
\dssectionheader{Preprocessing/cleaning/labeling}

\dsquestionex{Was any preprocessing/cleaning/labeling of the data done (e.g., discretization or bucketing, tokenization, part-of-speech tagging, SIFT feature extraction, removal of instances, processing of missing values)?}{If so, please provide a description. If not, you may skip the remainder of the questions in this section.}

\dsanswer{No.}

\dsquestionex{Was the “raw” data saved in addition to the preprocessed/cleaned/labeled data (e.g., to support unanticipated future uses)?}{If so, please provide a link or other access point to the “raw” data.}

\dsanswer{N/A}

\dsquestionex{Is the software used to preprocess/clean/label the instances available?}{If so, please provide a link or other access point.}

\dsanswer{N/A}

\dsquestion{Any other comments?}

\dsanswer{No.}

\bigskip
\dssectionheader{Uses}

\dsquestionex{Has the dataset been used for any tasks already?}{If so, please provide a description.}

\dsanswer{Yes, the data were used for face recognition bias tests in a survey project run by the dataset creators.}

\dsquestionex{Is there a repository that links to any or all papers or systems that use the dataset?}{If so, please provide a link or other access point.}

\dsanswer{No.}

\dsquestion{What (other) tasks could the dataset be used for?}

\dsanswer{This dataset could also be used in mitigating bias in facial recognition models as a training dataset. The facial categorization task could also utilize this dataset.}

\dsquestionex{Is there anything about the composition of the dataset or the way it was collected and preprocessed/cleaned/labeled that might impact future uses?}{For example, is there anything that a future user might need to know to avoid uses that could result in unfair treatment of individuals or groups (e.g., stereotyping, quality of service issues) or other undesirable harms (e.g., financial harms, legal risks) If so, please provide a description. Is there anything a future user could do to mitigate these undesirable harms?}

\dsanswer{No.}

\dsquestionex{Are there tasks for which the dataset should not be used?}{If so, please provide a description.}

\dsanswer{Facial recognition audits.}

\dsquestion{Any other comments?}

\dsanswer{No.}

\bigskip
\dssectionheader{Distribution}

\dsquestionex{Will the dataset be distributed to third parties outside of the entity (e.g., company, institution, organization) on behalf of which the dataset was created?}{If so, please provide a description.}

\dsanswer{Yes, the data will be shared publicly.}

\dsquestionex{How will the dataset will be distributed (e.g., tarball on website, API, GitHub)}{Does the dataset have a digital object identifier (DOI)?}

\dsanswer{GitHub.}

\dsquestion{When will the dataset be distributed?}

\dsanswer{2021.}

\dsquestionex{Will the dataset be distributed under a copyright or other intellectual property (IP) license, and/or under applicable terms of use (ToU)?}{If so, please describe this license and/or ToU, and provide a link or other access point to, or otherwise reproduce, any relevant licensing terms or ToU, as well as any fees associated with these restrictions.}

\dsanswer{No.}

\dsquestionex{Have any third parties imposed IP-based or other restrictions on the data associated with the instances?}{If so, please describe these restrictions, and provide a link or other access point to, or otherwise reproduce, any relevant licensing terms, as well as any fees associated with these restrictions.}

\dsanswer{No.}

\dsquestionex{Do any export controls or other regulatory restrictions apply to the dataset or to individual instances?}{If so, please describe these restrictions, and provide a link or other access point to, or otherwise reproduce, any supporting documentation.}

\dsanswer{No.}

\dsquestion{Any other comments?}

\dsanswer{No.}

\bigskip
\dssectionheader{Maintenance}

\dsquestion{Who will be supporting/hosting/maintaining the dataset?}

\dsanswer{This dataset will be hosted on GitHub and the authors of this paper will continue to support the dataset, performing any necessary maintenance.}

\dsquestion{How can the owner/curator/manager of the dataset be contacted (e.g., email address)?}

\dsanswer{\reviewing{\authorsident}{sdooley1@cs.umd.edu}}

\dsquestionex{Is there an erratum?}{If so, please provide a link or other access point.}

\dsanswer{A list of erratum is displayed and updated in the README of the project's GitHub.}

\dsquestionex{Will the dataset be updated (e.g., to correct labeling errors, add new instances, delete instances)?}{If so, please describe how often, by whom, and how updates will be communicated to users (e.g., mailing list, GitHub)?}

\dsanswer{Problematic images will be addressed when brought to attention, and an amended dataset will be released through GitHub. }

\dsquestionex{If the dataset relates to people, are there applicable limits on the retention of the data associated with the instances (e.g., were individuals in question told that their data would be retained for a fixed period of time and then deleted)?}{If so, please describe these limits and explain how they will be enforced.}

\dsanswer{Images and identities were curated from other well established datasets CelebA and Labeled Faces in the Wild. Please defer to the practices followed in said parent datasets.}

\dsquestionex{Will older versions of the dataset continue to be supported/hosted/maintained?}{If so, please describe how. If not, please describe how its obsolescence will be communicated to users.}

\dsanswer{All versions of the dataset will continue to be hosted on GitHub as different releases of the dataset.}

\dsquestionex{If others want to extend/augment/build on/contribute to the dataset, is there a mechanism for them to do so?}{If so, please provide a description. Will these contributions be validated/verified? If so, please describe how. If not, why not? Is there a process for communicating/distributing these contributions to other users? If so, please provide a description.}

\dsanswer{GitHub pull requests.}

\dsquestion{Any other comments?}

\dsanswer{No.}